\documentclass[letterpaper]{article} 
\usepackage{aaai2027}  
\usepackage[hyphens]{url}  
\usepackage{graphicx} 
\usepackage{natbib}  
\usepackage{caption} 
\usepackage{algorithm}
\usepackage{algorithmic}

\usepackage{newfloat}
\usepackage{listings}
\DeclareCaptionStyle{ruled}{labelfont=normalfont,labelsep=colon,strut=off} 
\floatstyle{ruled}
\newfloat{listing}{tb}{lst}{}
\floatname{listing}{Listing}

\usepackage{booktabs}

\usepackage{amsfonts}
\usepackage{tcolorbox}
\usepackage{multirow}
\usepackage{tcolorbox}
\usepackage{xspace}
\usepackage{amsmath}
\newcommand{\sys}{{MCTS-Report}\xspace}
\newcommand{\bench}{{MMRBench}\xspace}

\title{Monte Carlo Tree Search for Table-to-Multimodal Report Generation}
\author{
    Teng lin,
    Zhiyang Zhang, 
    Yuyu Luo,
    Nan Tang
}
\affiliations{
    The Hong Kong University of Science and Technology (Guangzhou)\\

tlin280@connect.hkust-gz.edu.cn,
\{yuyuluo,nantang\}@hkust-gz.edu.cn
}

\begin{document}

\maketitle

\begin{abstract}
Automatically generating professional multimodal reports comprising both textual analysis and visual charts from structured tabular data is a critical challenge in data intelligence. Existing methods suffer from fixed linear pipelines and isolated subtask processing, which hinder joint optimization of factual accuracy, visual quality, and narrative coherence. To address these issues, this paper proposes \sys, a Monte Carlo Tree Search (MCTS)-driven framework that formulates multimodal table-to-report generation as a progressive construction process over a structured search space. The core idea is to decompose report generation into atomic actions, including chapter planning, visualization task identification, chart generation, insight organization, and narrative refinement, each executed by an LLM based on dynamic reasoning conditioned on the current report state. We use an LLM to generate step-by-step reasoning and actions during MCTS, storing the reasoning trajectory in each node for context-aware, coherent report construction. To guide the search, we design a multi-dimensional reward function that jointly evaluates numerical fact consistency (via SQL), chart quality, chart-text alignment, and structural completeness, while incorporating a diversity penalty to suppress repeated charts and a precondition check to prune invalid actions. We also construct \bench, a comprehensive benchmark comprising real-world tables from six domains, paired with expert-refined reference report structures and verifiable key insights. Experiments on \bench demonstrate that \sys significantly outperforms strong baselines across structural completeness, numerical accuracy, chart-text alignment, and insight novelty, achieving a 77.9 overall score.
\end{abstract}

\section{Introduction}

Automatically generating professional multimodal reports containing both textual analysis and visual charts from structured tables is an increasingly important capability in data intelligence~\cite{huang2026mmdeepresearch,lin2026docsageinformationstructuringagent}. Unlike traditional table question answering or plain text summarization, multimodal report generation requires models to simultaneously parse tabular data, generate coherent natural language narratives, design and render appropriate charts, and finally produce a report that integrates text and visuals with logical coherence and factual trustworthiness. As enterprises and research institutions increasingly deploy AI agents for data analysis, enabling systems to reliably generate and evaluate such multimodal reports has become an urgent and challenging research problem~\cite{lin2025lightkggsimpleefficientknowledge, lin2026annoretrieveefficientstructuredretrieval}.

Despite recent progress in table reasoning and report generation, existing methods suffer from two fundamental limitations. \textbf{(1) Fixed linear pipelines.} Most current systems follow predetermined execution paths(e.g., table parsing → chart generation → text writing → polishing). This rigid sequencing prevents any form of backtracking or global optimization: once a chart is generated based on an incomplete understanding of the data, later textual analysis must adapt to the chart rather than the chart serving the analysis.\textbf{(2) Isolated subtask processing.} Existing methods process subtasks such as table parsing, chart generation, and content evaluation sequentially or in isolation. These objectives are optimized separately, without mutual verification or adjustment. For example, chart generation may prioritize visual aesthetics while ignoring factual fidelity to the source table, or textual analysis may make numerical claims that are not visually supported by the accompanying charts. Recent studies have identified chart-text inconsistency as a persistent issue in data-driven report generation~\cite{lin2026evidfuse,yang2025multimodal}, with staged pipelines often leading to ``insight freezing'' and shallow analysis~\cite{lin2026evidfuse,lin2025srag,lin2025Simplifying, lin2025Structured}. 

Beyond these methodological shortcomings, evaluation benchmarks also face critical gaps. Early benchmarks such as WikiTableQuestions~\cite{pasupat2015compositional}, FeTaQA~\cite{nan2022fetaqa}, and TabFact~\cite{chen2020tabfact} focus on narrow table question answering or sentence-level text generation. More recent benchmarks like T2R-Bench~\cite{zhang2025t2rbench} and DDR-Bench~\cite{liu2026huntinstead} support article-level report generation but remain confined to textual outputs, entirely ignoring visual charts, an essential component of business reports. Meanwhile, the LLM-as-a-judge paradigm is vulnerable to reward hacking~\cite{NEURIPS2023_91f18a12,zhao2026one}, and existing visualization evaluation methods (e.g., VisEval~\cite{chen2024viseval}, VIS-Shepherd~\cite{pan2025vis} disconnected from both the source tables and the accompanying textual analysis. MMDR‑Bench~\cite{huang2026mmdeepresearch} proposes multi‑stage evaluation with visual evidence fidelity checks for deep research reports.
No benchmark currently integrates structured tables, multimodal report generation, and verifiable multi-dimensional evaluation in a unified manner~\cite{lin-etal-2025-mebench}.

To address both the methodological and evaluation shortcomings, we propose \sys, a novel framework that fundamentally re-conceptualizes multimodal report generation. Instead of engineering a complex pipeline of specialized agents, we formulate report construction as a structured search problem over a tree-shaped space of partial reports. In this formulation, the root node represents an empty report framework, leaf nodes correspond to complete reports, and each edge denotes a construction action, such as planning a chapter, identifying a chart-worthy insight, generating a visualization, or refining narrative flow.

To navigate this exponentially large search space efficiently, we employ Monte Carlo Tree Search (MCTS)~\cite{mcts}. MCTS provides a principled mechanism for balancing exploration (trying novel report structures) and exploitation (refining promising ones) through iterative selection, expansion, simulation, and backpropagation. \sys uses a single LLM as the unified ``action-evaluation'' engine throughout the search process. 

This design offers three decisive advantages over existing methods. (1) MCTS enables global optimization: the system can backtrack from dead ends and explore alternative report structures, avoiding the ``insight freezing'' of linear pipelines. (2) The unified LLM-driven framework eliminates the need for fragile multi-agent coordination, simplifying system design and reducing error propagation. (3) The self-supervised reward ensures that factual correctness and cross-modal consistency are optimized jointly, without relying on potentially biased LLM-as-a-judge evaluations during training.

Alongside \sys, we construct \bench, a comprehensive benchmark for table-to-multimodal report generation. \bench comprises 185 real-world tables from six domains (finance, manufacturing, healthcare, education, retail, and IT operations), each paired with expert-refined reference reports and verifiable key insight points. The benchmark supports systematic evaluation across four dimensions: structural completeness, numerical accuracy, chart-text alignment, and insight novelty. Unlike prior benchmarks, \bench integrates structured tables, multimodal outputs, and multi-dimensional evaluation in a unified framework.

\textbf{Contributions.} This work makes three main contributions:

\begin{itemize}
    \item \textbf{\sys Framework:} Proposes a novel MCTS-driven framework that reformulates multimodal report generation as a structured search problem, using a single LLM as the unified action-evaluation engine to jointly optimize factual accuracy, visual quality, and narrative coherence.
    \item \textbf{\bench Benchmark:} Constructs a comprehensive benchmark with expert-curated reference reports and verifiable key insight points, supporting systematic evaluation of structural completeness, numerical accuracy, chart-text alignment, and insight novelty.
    \item \textbf{Extensive Experimental Validation:} \sys achieves significant improvements over 12 baselines across all metrics on \bench, with ablation studies confirming the critical contributions of MCTS planning, the unified LLM-driven action-evaluation mechanism, and self-supervised reward feedback.
\end{itemize}

\section{Related Work}

\begin{figure*}[t!]
\centering
\includegraphics[width=\linewidth]{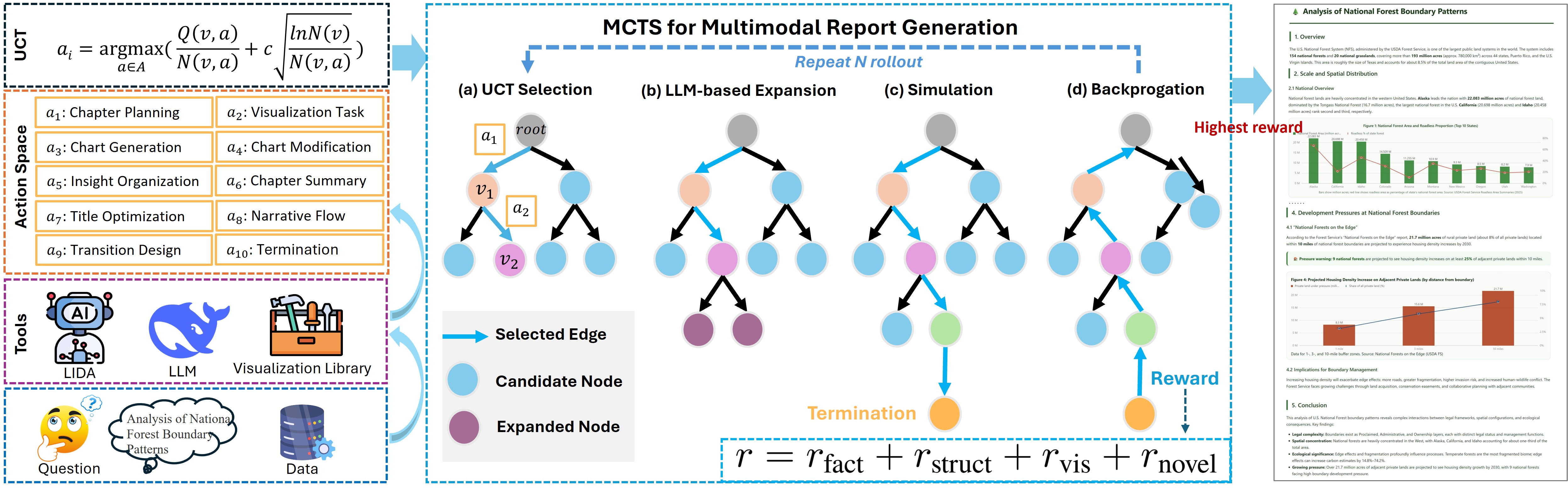}
\caption{The pipeline of \sys.}
\label{fig:pipeline}
\end{figure*}

\subsection{Benchmarks for Table-to-Report Generation}

Early table reasoning benchmarks mainly focused on narrow tasks. WikiTableQuestions~\cite{pasupat2015compositional}, FeTaQA~\cite{nan2022fetaqa}, and TabFact~\cite{chen2020tabfact} evaluate question answering over semi-structured tables. ToTTo~\cite{parikh2020totto} and HiTab~\cite{cheng2022hitab} address controlled table-to-text generation. However, these tasks only require sentence-level outputs and do not capture the complexity of generating article-level analytical reports.
To bridge this gap, T2R-Bench~\cite{zhang2025t2rbench} formulates the table-to-report task and constructs a benchmark of 457 real-world industrial tables spanning 19 domains. But it remains confined to textual outputs without visualizations. DDR-Bench~\cite{liu2026huntinstead} pushes toward autonomous data exploration: agents investigate structured databases without predefined questions and produce insight reports, evaluated via checklist-based fact verification. 
Several works have attempted to incorporate visual elements. Mul
timodalReportBench~\cite{yang2025multimodal}, introduced by Multimodal DeepResearcher, provides a benchmark for text-chart interleaved reports generated from web-based deep research. However, it does not target table-grounded report generation, and its evaluation relies on style and consistency metrics rather than numerical fact verification against source tables.

\subsection{Evaluation Frameworks for Multimodal Reports}

Evaluating multimodal reports is challenging due to the need to jointly assess textual and visual quality. The LLM-as-a-judge paradigm is widely adopted for scalability but suffers from fundamental vulnerabilities: recent work shows that reasoning judges can be deceived by adversarial outputs that achieve high rewards while violating task intent~\cite{liu2026examining}.
In the visualization domain, VisEval~\cite{chen2024viseval} and VIS-Shepherd~\cite{pan2025vis} employ LLMs to judge charts within NL2VIS pipelines. However, these approaches evaluate charts in isolation, disconnected from textual context and source tables. A unified evaluation framework that jointly measures factual grounding, visual fidelity, and narrative coherence in table-grounded multimodal reports remains absent.

\section{\sys Framework}
\label{sec:system}

\subsection{Task Formulation}

The multimodal report generation task is defined as follows: given one or more structured tables $T = \{T_1, T_2, \ldots, T_n\}$ and a natural language query $q$ specifying the analytical focus, a model must produce a multimodal report $\mathcal{R} = (R_{\text{text}}, R_{\text{vis}})$. The textual component $R_{\text{text}}$ is an article-length analytical report containing structured chapters (e.g., Overview, Data Overview, Core Analysis, Conclusions and Recommendations). The visual component $R_{\text{vis}}$ consists of one or more generated charts (e.g., bar charts, line plots, pie charts, scatter plots) embedded in the text, directly supporting the narrative with evidence from the tables.

\subsection{Framework Overview}

Unlike prior work that employs fixed linear pipelines or multi-agent coordination, \sys reformulates report generation as a \textbf{structured search problem} over a tree-shaped space. As illustrated in Figure~\ref{fig:pipeline}, the framework consists of four core components:

\begin{itemize}
    \item \textbf{Search Tree Representation}: Defines the state space, action space, and transition rules for report construction.
    \item \textbf{MCTS Search Engine}: Iteratively explores the tree through Selection, Expansion, Simulation, and Backpropagation.
    \item \textbf{LLM Execution Engine}: A single LLM is invoked at key MCTS stages to generate concrete content (actions, report drafts) based on context-specific prompts.
    \item \textbf{Self-Supervised Reward Function}: Evaluates completed report drafts via automated verification against source tables and structural rules, providing feedback signals for search guidance.
\end{itemize}

Crucially, the LLM serves as a unified reasoning engine throughout the process, rather than as multiple specialized agents with distinct identities and coordination protocols. All capabilities, including data parsing, chart generation, insight writing, and quality assessment, are realized through different prompting strategies applied to the same underlying LLM, conditioned on the current search state.


\subsection{Search Tree Formulation}

We represent the space of partial reports as a tree $\Psi = (V, E)$, where:

\begin{itemize}
    \item \textbf{Nodes $V$}: Each node $v \in V$ represents a partial report state. The root node $v_0$ contains the input tables $T$, the query $q$, and an empty report skeleton. Intermediate nodes store accumulated report content (completed chapters, generated charts, written insights) along with metadata such as the current chapter index and completed action history. A leaf/terminal node $v_t$ represents a complete report where all chapters are finalized.
    \item \textbf{Edges $E$}: Each edge $e \in E$ corresponds to the execution of a \textbf{construction action} $a \in \mathcal{A}$, transitioning from one partial state to the next.
    \item \textbf{Path to a Complete Report}: A complete report $\mathcal{R}$ corresponds to a path from $v_0$ to a terminal node $v_t$, where the sequence of actions composes the full generation trajectory.
\end{itemize}

\subsubsection{Action Space}

We define a comprehensive action space $\mathcal{A} = \{a_1, \ldots, a_7\}$ covering the full report construction lifecycle. Each action is associated with a specific prompt template that instructs the LLM on the expected output format. Table~\ref{tab:action_space} summarizes the action space.

\begin{table*}[t]
\centering
\begin{tabular}{cll}
\toprule
\textbf{Action} & \textbf{Name} & \textbf{Description}\\
\midrule
$a_1$ & Chapter Planning & Generate report chapter outline based on query $q$ and table schema \\
$a_2$ & Visualization Task & Identify analytical points worthy of chart support; specify chart type and data mapping \\
$a_3$ & Chart Generation & Generate executable code (Vega-Lite/Matplotlib) to render the chart  \\
$a_4$ & Chart Modification & Refine axis labels, colors, or legend based on self-critique \\
$a_5$ & Insight Organization & Write analytical paragraphs grounded in table data, linked to generated charts  \\
$a_6$ & Chapter Summary & Synthesize all insights within a chapter into a concise summary paragraph \\
$a_7$ & Title Optimization & Polish report title and section headings for clarity, impact, and consistency \\
$a_8$ & Narrative Flow & Improve transitions between paragraphs and chapters; polish section headings  \\
$a_9$ & Transition Design & Add bridging sentences between major sections to ensure narrative continuity\\
$a_{10}$ & Termination & Mark report as complete; no further actions \\
\bottomrule
\end{tabular}
\caption{Action Space for \sys}
\label{tab:action_space}
\end{table*}

\subsubsection{Action Ordering Constraints}

Not all action sequences are valid. To ensure logical coherence, we impose a \textbf{transition constraint matrix} (Table~\ref{tab:transitions}). 

\begin{table}[ht]
\centering
\begin{tabular}{ll}
\toprule
\textbf{Current Action} & \textbf{Valid Next Actions} \\
\midrule
- (Root) & $a_1$ \\
$a_1$ (Chapter Planning) & $a_2, a_5$ \\
$a_2$ (Viz Task ID) & $a_3$ \\
$a_3$ (Chart Generation) & $a_4, a_5$ \\
$a_4$ (Chart Modification) & $a_5$ \\
$a_5$ (Insight Organization) & $a_2, a_5, a_6, a_8$ \\
$a_6$ (Chapter Summary) & $a_7, a_8, a_9$ \\
$a_7$ (Title Optimization) & $a_{10}$ \\
$a_8$ (Narrative Flow) & $a_7, a_9, a_{10}$ \\
$a_9$ (Transition Design) & $a_7, a_{10}$ \\
$a_{10}$ (Termination) & -- \\
\bottomrule
\end{tabular}
\caption{Valid Action Transitions}
\label{tab:transitions}
\end{table}

\subsection{MCTS-Driven Report Generation}

\sys performs $N_{\text{rollout}}$ independent MCTS rollouts per query. Each rollout consists of four phases, with the LLM playing distinct roles in Expansion, Simulation, and Backpropagation, but not in Selection.

\subsubsection{Phase 1: Selection (UCT-Based)}

Starting from the root node, the algorithm traverses the tree by iteratively selecting child nodes. For a given node $v$ with action set $\mathcal{A}(v)$ (determined by Table~\ref{tab:transitions}), the selection follows a two-step priority:

\begin{itemize}
    \item \textbf{Breadth-first priority}: If any child node corresponding to an action $a \in \mathcal{A}(v)$ has never been visited ($N(v,a) = 0$), select that action preferentially (random tie-breaking). This ensures comprehensive exploration of all possible action types early in the search.
    \item \textbf{UCT-based selection}: If all actions from $v$ have been visited at least once, compute the UCT score for each action and select the maximum:
    \begin{equation}
        \mathrm{UCT}(v, a) = \frac{Q(v, a)}{N(v, a)} + c \cdot \sqrt{\frac{\ln N(v)}{N(v, a)}},
    \end{equation}
    where $Q(v, a)$ is the cumulative reward obtained from paths that took action $a$ from node $v$, $N(v, a)$ is the visit count of action $a$ from node $v$, $N(v)$ is the total visit count of node $v$, and $c = 1.414$ is the exploration constant.
\end{itemize}

The traversal continues until reaching a node that is either terminal or has unexpanded children (i.e., not all legal actions from this node have been generated as child nodes).


\subsubsection{Phase 2: Expansion (LLM-based Action Generation)}

When the Selection phase reaches a non-terminal node $v$ with unexplored actions, the Expansion phase generates new child nodes. For each legal action $a \in \mathcal{A}_{\text{valid}}(v)$ that has not yet been expanded from $v$:

\begin{itemize}
    \item Retrieve the action-specific prompt template from a prompt library.
    \item Construct the full prompt by injecting:
    \begin{itemize}
        \item The original query $q$ and table schema (summarized from the root node).
        \item The accumulated reasoning trajectory (actions and their outputs from the current path).
        \item Metadata about already-generated charts and insights.
    \end{itemize}
    \item Invoke the LLM $M$ to generate the action output. We sample $k = 3$ times with temperature $T_{\text{expansion}} = 0.8$ to encourage diversity.
    \item For each generated output, create a new child node $v'$ representing the updated partial report state.
\end{itemize}

The newly created child nodes are added to the tree with initial statistics $N = 0, Q = 0$.

\subsubsection{Phase 3: Simulation (LLM-Driven Rapid Drafting)}

From the newly expanded node (or, in some implementations, from a randomly selected child), the Simulation phase generates a complete report draft to evaluate the path's potential. This phase proceeds as follows:

\begin{itemize}
    \item Starting from the current partial state, repeatedly select and execute actions according to a \textbf{fast rollout policy}. We use a lightweight policy: at each step, randomly select a legal action from the transition matrix (Table~\ref{tab:transitions}) with uniform probability, but favor actions that lead toward termination (e.g., $a_5$ and $a_6$).
    \item For each selected action, invoke the LLM with the corresponding prompt template to generate content. To reduce computational cost during simulation, we use \textbf{lower sampling temperature} ($T_{\text{sim}} = 0.3$) and \textbf{shorter prompts} (omitting verbose reasoning instructions).
    \item Continue until the Termination action $(a_{10})$ is reached, yielding a complete report draft $\mathcal{R}_{\text{draft}}$.
\end{itemize}

The simulation path is not permanently added to the search tree; it is a temporary exploration used only for reward evaluation. However, the actions taken during simulation are not persisted to avoid bloating the tree with low-quality paths.

\subsubsection{Phase 4: Backpropagation (Self-Supervised Reward + Statistics Update)}

After obtaining $\mathcal{R}_{\text{draft}}$, we compute a self-supervised reward $r$ that evaluates the draft without requiring external judge models or human annotations.

\paragraph{Reward Function}
The reward is a weighted sum of four sub-scores:
\begin{equation}
    r = r_{\text{fact}} + r_{\text{struct}} + r_{\text{vis}} + r_{\text{novel}},
\end{equation}

\begin{itemize}
    \item \textbf{$r_{\text{fact}}$} (Factual Accuracy): Extract all numerical claims from $\mathcal{R}_{\text{draft}}$ (e.g., ``sales increased by 15\%'').For each claim, generate a verification SQL query against the source tables $T$. Execute the query and compare the result with the claim. A claim is correct if it matches within a 1\% tolerance. $r_{\text{fact}} = \frac{\#\text{correct claims}}{\#\text{total claims}}$.
    \item \textbf{$r_{\text{struct}}$} (Structural Completeness): Check whether $\mathcal{R}_{\text{draft}}$ contains all required chapters (Overview, Data Overview, Core Analysis, Conclusions). Check whether each chapter contains at least one paragraph. $r_{\text{struct}} = 1$ if both conditions are satisfied, otherwise 0.
    \item \textbf{$r_{\text{vis}}$} (Visual Quality): For each chart in $\mathcal{R}_{\text{draft}}$:
        \begin{itemize}
            \item Technical correctness: Chart code executes without errors; axes and legends are non-empty; title exists.
            \item Data fidelity: Extract rendered data points via OCR + parsing, compare with source data; tolerance $\le 1\%$.
        \end{itemize}
         $r_{\text{vis}} = \frac{1}{|C|} \sum_{c \in C} \mathbb{1}[\text{chart } c \text{ passes all checks}]$, where $C$ is the set of charts in the report.
    \item \textbf{$r_{\text{novel}}$} (Insight Novelty): Compare the report's key insight points against a set of common/trivial patterns (e.g., ``total sales increased''). Compute the proportion of insights that are not simple paraphrases of basic aggregations. We implement this via cosine similarity with a template bank; insights with similarity $< 0.7$ are considered novel.

\end{itemize}

\paragraph{Backpropagation}
After computing $r$, we traverse the path from the terminal node back to the root. For each node $u$ along the path and the action $a_u$ taken from it:
\begin{align}
    N(u, a_u) &\leftarrow N(u, a_u) + 1, \\
    Q(u, a_u) &\leftarrow Q(u, a_u) + r, \\
    N(u) &\leftarrow N(u) + 1.
\end{align}
These updated statistics influence future Selection phases through the UCT formula.

\subsection{Complexity and Efficiency Considerations}

The search space size for report generation is combinatorial: with 10 action types and up to 5 chapters per report, the number of possible paths grows exponentially. However, \sys achieves practical efficiency through:

\begin{itemize}
    \item \textbf{Action ordering constraints} (Table~\ref{tab:transitions}) pruning invalid paths early.
    \item \textbf{Deduplication} during expansion reducing redundant branches.
    \item \textbf{Simulation-phase cost control} using lower temperature and shorter prompts.
    \item \textbf{Early termination}: Simulation stops as soon as a path's partial reward falls below a dynamic threshold (implemented as an optional optimization).
\end{itemize}




\section{\bench Construction}
\label{sec:evaluation}

\subsection{Data Collection and Domain Coverage}

We collected 185 real-world tables from public sources and open-source tabular datasets~\cite{zhang2025t2rbench}, covering six domains: finance, manufacturing, healthcare, education, retail, and IT operations. Among them, 79 tables are in English and 131 are in Chinese, ensuring the benchmark’s capability for bilingual evaluation. All tables were manually reviewed to remove personally identifiable information and ensure sufficient depth for analysis. Based on these tables, we generated 386 task queries, of which 279 are single‑table tasks and 107 are multi‑table tasks. The inclusion of multi‑table tasks increases the difficulty of evaluating cross‑table reasoning and complex data association capabilities. Key statistics of \bench are summarized in Table~\ref{tab:stats}.

\begin{table}[htbp]
\centering
\small
\begin{tabular}{lc}
\toprule
Property & Value \\
\midrule
Number of tables & 185 \\
English / Chinese tables & 79 / 131\\
Single‑table / Multi‑table tasks & 279 / 107\\
Average cells per table & 420,000 \\
Number of domains & 6 \\
Extremely large tables (\(>50\)K cells) & 38 \\
Average tasks per table & 1.84 \\
Complex header tables (\%) & 32.4\% \\
Average reference keypoints per task & 4.75 \\
Total reference keypoints & 1,834 \\
\bottomrule
\end{tabular}
\caption{Key statistics of \bench}
\label{tab:stats}
\end{table}

\subsection{Query Generation}

We used a template-based self-instruct approach to generate analytical queries. Ten templates covering common business reporting scenarios (trend analysis, comparative evaluation, ranking, anomaly detection, composition analysis, correlation exploration, forecasting, segment analysis, year-over-year change, cross-dimensional interaction) were populated with table-specific metadata and expanded into 3–5 concrete query candidates by GPT-4o. Two independent domain experts and one senior expert then selected queries based on three criteria: (1) \textbf{answerability}: fully solvable from the provided tables without external knowledge; (2) \textbf{analytical focus}: targeting a single coherent narrative thread; (3) \textbf{complementarity}: distinct non-overlapping analytical dimensions for multiple queries per table. This yielded 386 high-quality task queries.

\subsection{Reference Report and Keypoint Annotation}

Reference reports were generated via a two-stage “LLM generation + human refinement” pipeline.

\textbf{Stage 1: LLM generation of candidate reports.} We used GPT-4o (temperature 0.2) to generate three independent initial reports per query. To encourage diversity, the three reports used slightly different prompts emphasizing numerical-driven, logic-driven, or narrative-driven analysis. Each report was about 800–1200 words, containing 3–5 chapters and 4–8 chart placeholders.

\textbf{Stage 2: Human refinement and keypoint extraction.} Two annotators (with data analysis and report writing experience) performed the following on the three initial reports: (1) Selected the best candidate as the base.
    (2) Corrected numerical errors: verified all numerical claims against source tables using SQL and fixed mismatches.
    (3) Standardized chapter structure: unified to the “Overview → Dimension Analysis → Conclusions and Recommendations” framework.
    (4) Extracted keypoints: distilled 5–10 non-trivial factual statements from the report. Keypoints had to be strictly derivable from the tables, containing no subjective evaluations.

Disagreements were resolved by a third senior expert. The final output averaged 4.75 keypoints per task (1,834 total). A reproducibility audit on 50 randomly selected tasks by three independent annotators achieved Fleiss' \(\kappa = 0.85\), confirming the reliability of the annotation process.

\subsection{Evaluation Protocol}


\textbf{Evaluation method.} The evaluation protocol in \bench follows a strict separation principle: the self‑supervised reward used internally within \sys’s MCTS process (to guide search) and the benchmark evaluation (for comparing methods) are independent of each other. The latter uses a separate single‑model LLM‑as‑judge framework that scores reports on a 1–100 scale across four dimensions: structural completeness, numerical accuracy, chart‑text alignment, and insight novelty. The single‑model evaluation avoids potential reward hacking loops with the MCTS reward model. 
\section{Experiments}

\subsection{Experimental Setup}

\textbf{Models.} We evaluate \sys on \bench against a wide range of state-of-the-art models across three categories: (i) vision-language models with native chart generation (GPT-4o~\cite{achiam2023gpt}, GPT-4.1, Gemini-2.5-Pro, Gemini-3.5-Flash, Claude-4.5-Sonnet~\cite{anthropic2025claudesonnet45}, Qwen3-VL-235B~\cite{qwen3-vl-235b}); (ii) code-augmented multimodal systems (DeepSeek-R1 with code interpreter~\cite{guo2025deepseek}, Qwen3-Coder-32B, TableGPT2-7B~\cite{su2024tablegpt2}); and (iii) deep research agents (Gemini Deep Research~\cite{google-gemini-deep-research}, ChatGPT Deep Research, Perplexity Sonar Deep Research). 

\textbf{Evaluation Protocol.} Each model receives a task instance consisting of one or more tables and an analytical instruction $q$, and generates a multimodal report with both text and embedded visualizations. All models use the same prompt template, with a maximum of 4096 tokens for the textual report and up to 8 visualizations.

\textbf{Evaluation Metrics.} Following \bench's evaluation protocol, we use GPT-4o (temperature 0.2) as a single judge model, Overall score is the average of four dimensions.

\textbf{Human Baseline.} Six professional business analysts independently generated reference reports on a stratified sample of 50 tasks following the same protocol. 

\begin{table*}[t]
\centering
\begin{tabular}{lcccc c}
\toprule
\textbf{Model} & \textbf{Structural} & \textbf{Numerical} & \textbf{Chart-Text} & \textbf{Novelty} & \textbf{Overall} \\
\midrule
\multicolumn{6}{c}{\textit{Vision-Language Models}} \\
GPT-4o & 75.4 & 55.8 & 68.5 & 34.2 & 58.5 \\
GPT-5 & 76.3 & 53.1 & 69.1 & 31.3 & 57.4 \\
Gemini-2.5-Pro & 76.1 & 57.3 & 71.2 & 35.5 & 60.0 \\
Gemini-3.5-Flash & 74.1 & 52.9 & 72.6 & 39.5 & 59.8 \\
Claude-4.5-Sonnet & 74.2 & 54.8 & 70.5 & 33.6 & 58.3 \\
Qwen3-VL-235B & 71.6 & 51.9 & 66.2 & 30.4 & 55.0 \\
\midrule
\multicolumn{6}{c}{\textit{Code-Augmented Multimodal Systems}} \\
DeepSeek-R1 & 78.2 & 59.4 & 74.5 & 38.7 & 62.7 \\
Qwen3-Coder-32B & 72.9 & 52.4 & 69.8 & 31.5 & 56.7 \\
TableGPT2-7B & 65.3 & 44.7 & 60.2 & 23.2 & 48.4 \\
\midrule
\multicolumn{6}{c}{\textit{Deep Research Agents}} \\
Gemini Deep Research & 77.5 & 58.1 & 73.8 & 37.2 & 61.7 \\
ChatGPT Deep Research & 75.2 & 55.9 & 71.5 & 35.0 & 59.4 \\
Perplexity Sonar DR & 73.8 & 53.5 & 69.2 & 33.8 & 57.6 \\
\midrule
\multicolumn{6}{c}{\textit{Our Method}} \\
\textbf{\sys (DeepSeek-R1)} & \textbf{88.6} & \textbf{73.1} & \textbf{88.7} & \textbf{61.3} & \textbf{77.9} \\
\sys (GPT-4o) & 88.2 & 72.7 & 85.5 & 50.1 & 74.1 \\
\sys (Gemini-3.5-Flash) & 85.5 &67.4 &82.1 &52.1 &71.8 \\
\midrule
\multicolumn{6}{c}{\textit{Human Baseline}} \\
Human Expert & 94.8 & 91.2 & 89.5 & 88.5 & 91 \\
\bottomrule
\end{tabular}
\caption{Overall performance on \bench. Best results are bolded.}
\label{tab:main_results}
\end{table*}

\subsection{Main Results}

Table~\ref{tab:main_results} presents the overall and dimension-level performance of all evaluated models on \bench. We highlight the following key findings:

\textbf{Finding 1: MCTS-guided search consistently boosts performance across all base models.} \sys significantly improves every base model it is applied to. It is confirmed that search-based planning is particularly effective at improving factual reliability and cross-modal coordination, capabilities that are often weak in direct generation.

\textbf{Finding 2: \sys (DeepSeek-R1) establishes a new state-of-the-art, outperforming all baselines and rivaling human performance in chart-text alignment.} Overall score is 77.9, far exceeding the best baseline of DeepSeek-R1 (62.7, +15.2). The alignment of text and images is 88.7, close to the human score of 89.5, but there are still significant differences in novelty and numerical accuracy.

\textbf{Finding 3: Novelty remains a common bottleneck.} Although all models' novelty scores are far below the human baseline (\sys's 61.3 vs.~human 88.5), \sys achieves relative superiority through the exploration mechanism inherent in MCTS, which encourages the search to try less obvious analytical angles. 


\subsection{Ablation Studies}

To investigate the contribution of each component in \sys, we design four variants: \textbf{Variant A (w/o MCTS):} Directly prompts the underlying LLM with the same action space but without MCTS search, essentially performing a single rollout. \textbf{Variant B (w/o Self-Supervised Reward):} Uses random selection during search instead of UCT-guided selection; rewards are not backpropagated. \textbf{Variant C (Reduced Rollouts):} Limits MCTS to $N_{\text{rollout}} = 5$ instead of 10. \textbf{Variant D (Full \sys):} The complete framework. All variants use Deepseek-R1.

\begin{figure}[t!]
\centering
\includegraphics[width=\linewidth]{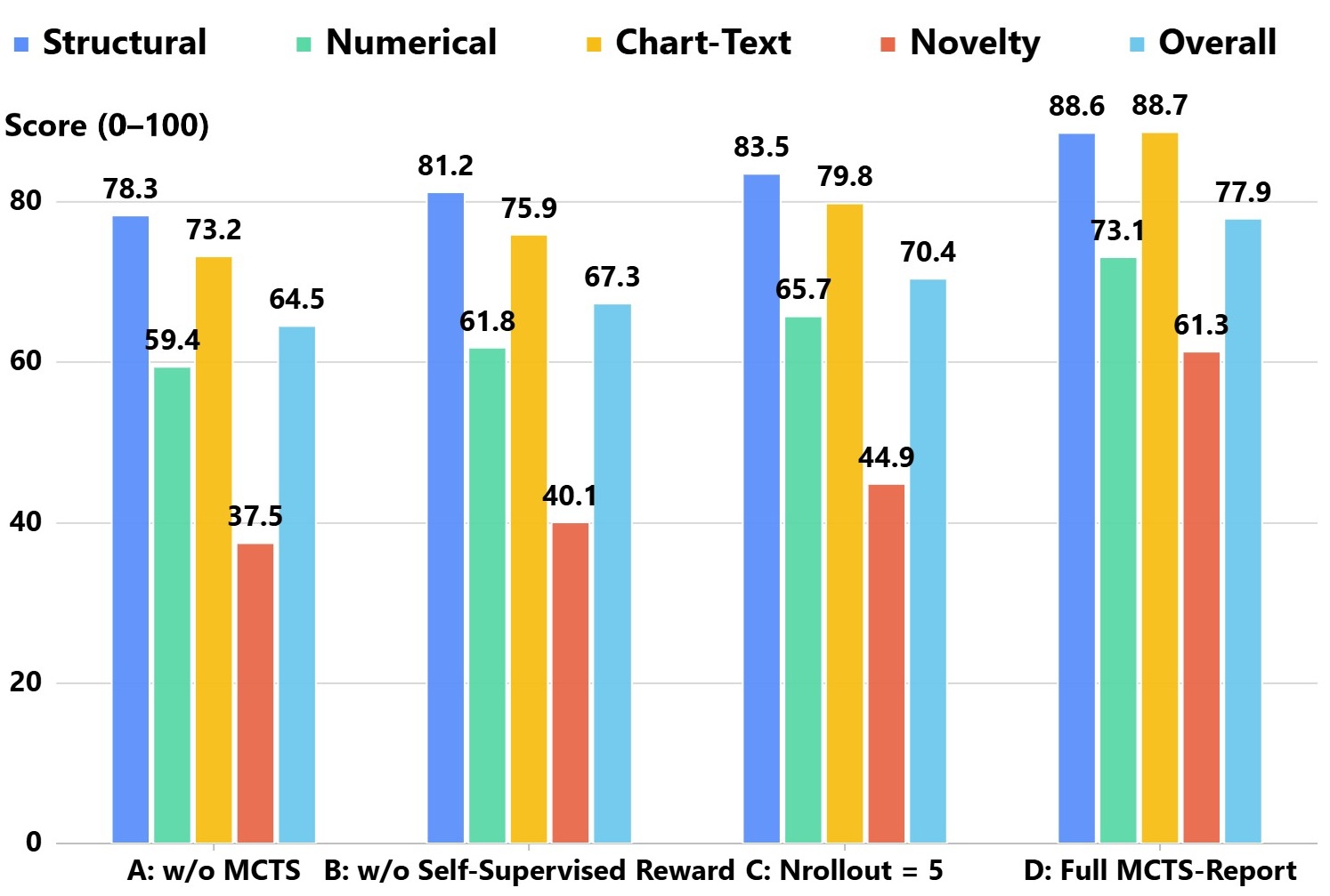}
\caption{Ablation results of \sys and its variants.}
\label{fig:ablation}
\end{figure}

Figure~\ref{fig:ablation} reports the performance of each variant. We observe:
\begin{itemize}
\item \textbf{MCTS planning is essential.} Variant A underperforms full \sys by 13.4 points overall, with the largest drops in numerical accuracy (-13.7) and novelty (-23.8). This confirms that the iterative search and backpropagation mechanism is critical for generating factually reliable and insightful reports.
\item \textbf{Self-supervised reward guidance matters.} Variant B lags behind full \sys by 10.6 points, with numerical accuracy suffering the most. Without reward feedback, the search cannot distinguish between promising and poor paths, reverting to near-random exploration.
\item \textbf{More rollouts improve performance.} Variant C (5 rollouts) achieves 70.4 overall, while full \sys reaches 77.9. This positive correlation suggests that deeper exploration of the search space yields better reports, though with diminishing returns.
\end{itemize}

\subsection{Error Analysis}

\begin{figure}[t!]
\centering
\includegraphics[width=\linewidth]{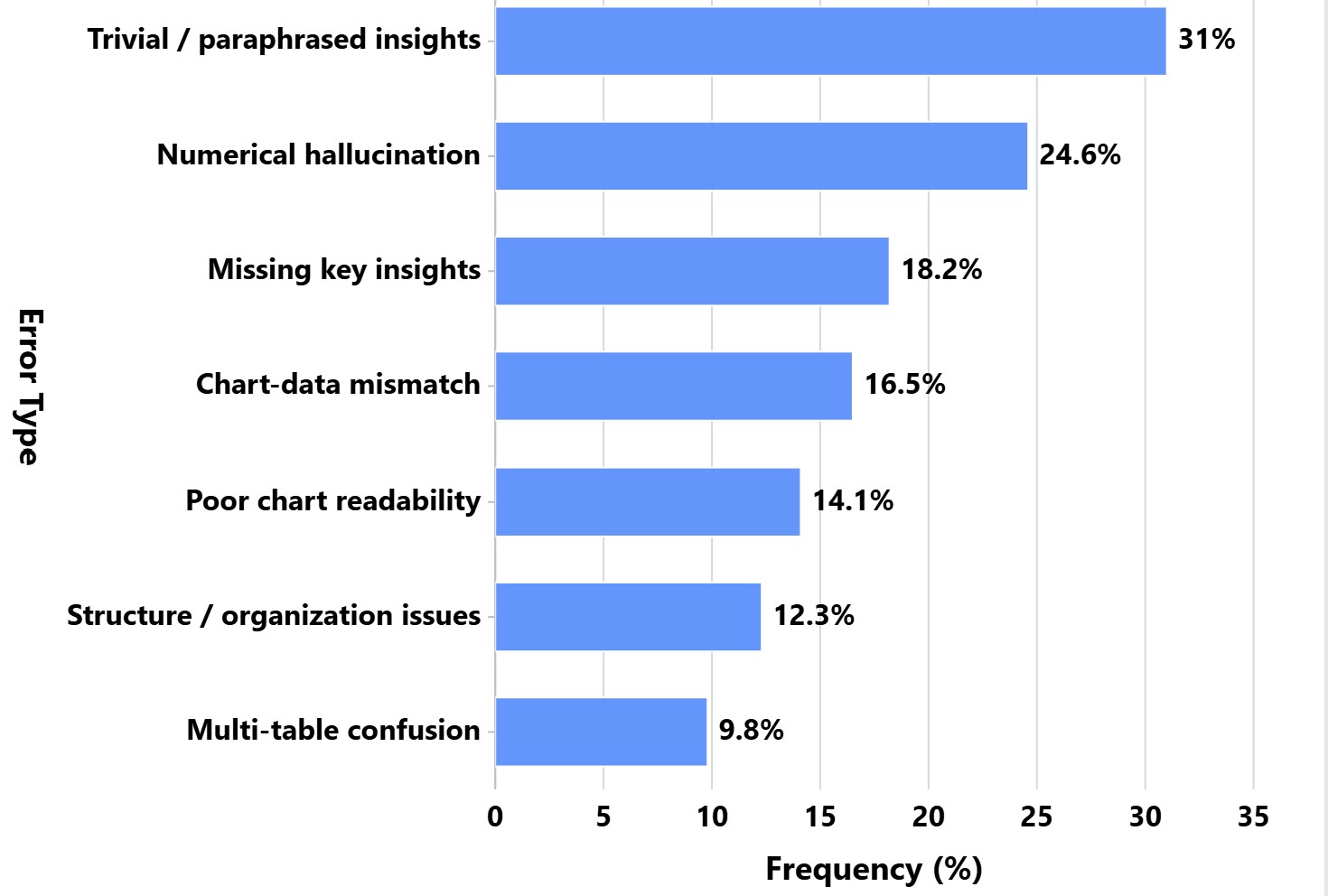}
\caption{Distribution of primary error types in generated reports. A single report may contain multiple errors.}
\label{fig:error}
\end{figure}

We manually analyze 200 randomly sampled reports generated by \sys to categorize failure modes. Figure~\ref{fig:error} presents the distribution of primary error types. \textbf{Numerical hallucination (24.6\%)} is substantially lower than that of the best baseline DeepSeek-R1 (38.5\%), thanks to the self-supervised reward's SQL-based verification during search. \textbf{Trivial/paraphrased insights (31.0\%)} remain the most frequent error, consistent with the low novelty scores observed in Table~\ref{tab:main_results}, this suggests that while MCTS helps explore different structures, the underlying LLM's analytical depth remains a bottleneck. \textbf{Multi-table confusion (9.8\%)} is especially prevalent in finance and healthcare reports, where schemas are complex and joins are non-trivial. This calls for stronger cross-table reasoning mechanisms, potentially through more fine-grained reward signals during search.

\section{Conclusion}

This paper presented \sys, an MCTS-enhanced multi-agent framework for multimodal report generation from tables, and \bench, a comprehensive benchmark for systematic evaluation. Extensive experiments on \bench show that \sys achieves an overall score of 77.9, significantly outperforming other methods. Ablation studies confirm the critical contributions of MCTS planning, reward guidance, and search depth. Future work will focus on efficient search strategies, finer-grained reward models, reducing hallucinations through explicit numerical verification, enhancing novelty via retrieval-augmented generation or curiosity-driven exploration, extending the framework to interactive, user-in-the-loop report generation, and expanding \bench to more domains..

\bibliography{refs/custom}

\end{document}